\documentclass{article}

\usepackage[preprint]{neurips_2026}

\usepackage[utf8]{inputenc} 
\usepackage[T1]{fontenc}    
\usepackage{hyperref}       
\usepackage{url}            
\usepackage{booktabs}       
\usepackage{amsfonts}       
\usepackage{amsmath}        
\usepackage{mathtools}      
\usepackage{nicefrac}       
\usepackage{microtype}      
\usepackage{xcolor}         
\usepackage{graphicx}       
\usepackage{multirow}       
\usepackage{enumitem}       
\usepackage{wasysym}
\usepackage{tikz}
\usetikzlibrary{positioning, arrows.meta}

\newtheorem{proposition}{Proposition}

\newtheorem{definition}{Definition}

\newcommand{\chmark}{\checkmark}

\definecolor{detectblue}{HTML}{537DA3}
\definecolor{enforceteal}{HTML}{7D9565}
\definecolor{recoverorange}{HTML}{D68043}
\definecolor{dimAR}{HTML}{D8DED0}    
\definecolor{dimLC}{HTML}{F7DEC8}    
\definecolor{dimPC}{HTML}{D9E4E8}    
\definecolor{dimRA}{HTML}{D9E4E8}    
\definecolor{dimEI}{HTML}{D9E4E8}    
\definecolor{basegray}{HTML}{BDBDBD} 

\newcommand{\ScanProjects}{6}
\newcommand{\TotalFindings}{617}
\newcommand{\CriticalFindings}{269}
\newcommand{\CriticalPct}{44}
\newcommand{\BlockTier}{134}

\newcommand{\ToolMisusePct}{64}
\newcommand{\SkillCount}{18{,}899}

\newcommand{\AttackCases}{48}
\newcommand{\BenignCalls}{500}
\newcommand{\LatencyTrials}{1{,}000}
\newcommand{\FalsePositiveRate}{1.2}
\newcommand{\MedianLatency}{8.3\,ms}

\newcommand{\RecoveryIoU}{0.814--0.843}
\newcommand{\RecoveryTokenAcc}{0.818--0.867}
\newcommand{\RecoveryEdgeSim}{0.912--0.949}

\usepackage{xcolor}
\makeatletter
\@ifundefinedcolor{uscred}{\definecolor{uscred}{HTML}{990000}}{}
\makeatother

\usepackage[normalem]{ulem}

\newif\ifshowcomments
\showcommentsfalse   

\usepackage{xcolor}

\newif\ifshowstudentnotes
\showstudentnotesfalse   

\definecolor{mustred}{RGB}{160,20,20}
\definecolor{shouldblue}{RGB}{25,70,160}
\definecolor{protocolgreen}{RGB}{20,110,70}

\title{Auditable Agents}

\author{%
  Yi Nian$^{1,*}$ \quad
  Aojie Yuan$^{1,*}$ \quad
  Haiyue Zhang$^{1,*}$ \quad
  Jiate Li$^{1,*}$ \quad
  Li Li$^{1}$ \\
  \bf Xiyang Hu$^{2}$ \quad
  Hua Wei$^{2}$ \quad
  Xiongye Xiao$^{3}$ \quad
  Chaowei Xiao$^{4}$ \quad
  Yue Zhao$^{1}$ \\[0.5em]
  $^{1}$University of Southern California \quad
  $^{2}$Arizona State University \\[0.2em]
  $^{3}$University of Tennessee, Knoxville \quad
  $^{4}$Johns Hopkins University \\[0.3em]
  \texttt{\{yinian, aojieyua, haiyuez, jiateli, li.li02, yue.z\}@usc.edu} \\
  \texttt{\{xiyanghu, hua.wei\}@asu.edu} \quad
  \texttt{xxiao9@utk.edu} \quad
  \texttt{chaoweixiao@jhu.edu} \\[0.3em]
  {\small $^{*}$Equal contribution.}
}

\makeatletter
\renewcommand{\@noticestring}{%
  Preprint. A condensed version of this paper appears in the Proceedings of the
  ACM AI Leadership Summit 2026 (Visionary Papers track).%
}
\makeatother

\begin{document}

\maketitle

\begin{abstract}
LLM agents call tools, query databases, delegate tasks, and trigger external
side effects.
Once an agent system can act in the world, the question is no longer only
whether harmful actions can be prevented---it is whether those actions remain
\emph{answerable} after deployment.
We distinguish \emph{accountability} (the ability to determine compliance and
assign responsibility), \emph{auditability} (the system property that makes
accountability possible), and \emph{auditing} (the process of reconstructing
behavior from trustworthy evidence).
Our claim is direct: \textbf{no agent system can be accountable without
auditability}.

To make this operational, we define five dimensions of agent
auditability, i.e., action recoverability, lifecycle coverage, policy
checkability, responsibility attribution, and evidence integrity, and
identify three mechanism classes (\emph{detect}, \emph{enforce},
\emph{recover}) whose temporal information-and-intervention constraints
explain why, in practice, no single approach suffices.
We support the position with layered evidence rather than a single benchmark:
lower-bound ecosystem measurements suggest that even basic security
prerequisites for auditability are widely unmet (\TotalFindings\ security
findings across six prominent open-source projects); runtime feasibility results show that
pre-execution mediation with tamper-evident records adds only \MedianLatency\
median overhead; and controlled recovery experiments show that
responsibility-relevant information can be partially recovered even when
conventional logs are missing.
We propose an \emph{Auditability Card} for agent systems and identify six
open research problems organized by mechanism class.
\end{abstract}

\section{Introduction}
\label{sec:intro}

LLM agents do not only generate text. They delete files, send messages,
issue payments, invoke third-party skills, and cross permission
boundaries~\citep{yao2023react,qin2023toolllm,liu2025make}.
Once an agent system can cause external side effects, its failures are no
longer only content problems. They are system problems, and system
problems require a different safety guarantee than alignment or
pre-deployment evaluation can provide on their own.

Consider a deployed enterprise agent. It reads customer records, queries an
external API, drafts an email, and sends it on behalf of a human operator.
The next day, the recipient reports that the email disclosed information it
should not have. Three questions then become central:
\begin{quote}
\emph{What happened? Did the system comply with policy?
Who or what was responsible?}
\end{quote}
In most deployed agent systems today, these questions cannot be answered with
confidence. Existing observability tools provide limited support for
agent-specific intermediate steps, traces, tool calls, and
artifacts~\citep{dong2024agentops}. Error paths, retries, fallbacks, approvals,
and inter-agent handoffs are often missing or weakly
represented~\cite{barke2026agentrx}. Skill provenance is shallow~\cite{xu2026agent}. Even when records exist, they
rarely support mechanical policy checking, and they are seldom protected
against silent modification.

Recent work has made real progress on alignment, adversarial evaluation, and
runtime defenses~\citep{zou2023universal,qi2024finetuning,mazeika2024harmbench,xu2024safedecoding,inan2023llamaguard}.
These efforts reduce the probability of harmful actions. But they do not
answer the post-deployment question: once an agent system has acted, can its
behavior be reconstructed, checked against policy, and attributed to a
responsible component?

\paragraph{Position.}
\underline{\textbf{Agent systems should be auditable.}}
No agent system can be accountable without auditability. This paper makes
that claim precise and argues that auditability should be treated as a
first-class design and evaluation requirement for agent systems.

\paragraph{Three levels.}
We distinguish three related concepts. \textbf{Accountability} is the goal:
an auditor can determine whether the system complied with policy and assign
responsibility for violations. \textbf{Auditability} is the enabling system
property: the system produces, preserves, and exposes enough trustworthy
evidence to make accountability possible. \textbf{Auditing} is the process:
reconstructing behavior, checking policy, and assigning responsibility from
the available evidence. These three levels are distinct from observability,
monitoring, and alignment, which address related but different questions. We
return to these distinctions in \S\ref{sec:alternatives}.

\paragraph{The paper's contribution.}
This paper does not propose a single new algorithm. Instead, it advances a
systems position: \emph{agent auditability should be a first-class design and
evaluation target}. We make the following contributions:

\begin{itemize}[leftmargin=*]
    \item \textbf{Five-Dimensional Auditability Framework.} We define five conditions that are jointly necessary for a defensible post-deployment audit: action recoverability, lifecycle coverage, policy checkability, responsibility attribution, and evidence integrity (\S\ref{sec:dimensions}).

    \item \textbf{Mechanism Classes.} We identify three classes of mechanism (\emph{detect}, \emph{enforce}, and \emph{recover}) that operationalize the five dimensions across the system lifecycle, and argue that in practice no single temporal vantage point can satisfy all five (\S\ref{sec:lifecycle}).

    \item \textbf{Layered Evidence.} We support the position with three mutually reinforcing evidence blocks: public ecosystem measurements, runtime feasibility results, and missing-log recovery experiments (\S\ref{sec:evidence}).

    \item \textbf{Auditability Card and Open Problems.} We propose an Auditability Card for agent systems, a compact reporting artifact analogous to model cards~\citep{mitchell2019model}, and identify six open research problems organized by mechanism class (\S\ref{sec:agenda}).
\end{itemize}

\begin{figure*}[!t]
\centering
\includegraphics[width=\textwidth]{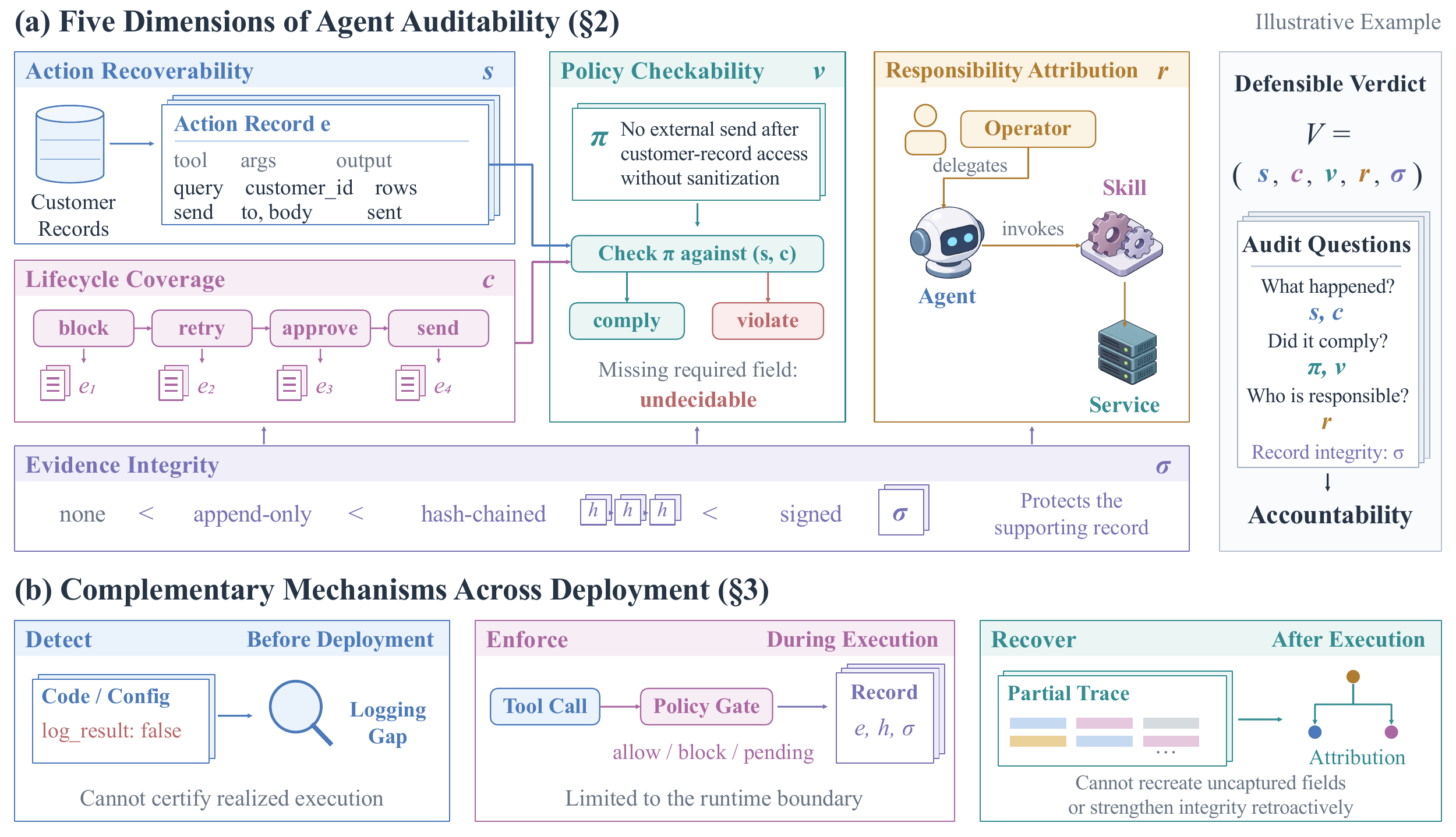}
\caption{Auditability dimensions and complementary mechanisms. (a) The five dimensions (\S\ref{sec:dimensions}) support a defensible audit verdict (Eq.~\ref{eq:verdict}). Action and lifecycle evidence support policy checking; a missing required field makes the assessment undecidable. Integrity protects the supporting record, with increasing assurance levels shown. (b) Detect, Enforce, and Recover (\S\ref{sec:lifecycle}) inspect or produce evidence at different stages and retain distinct limitations. Their dimension-level coverage appears in Table~\ref{tab:lifecycle-coverage}. Here, $e$ denotes an event record, $h$ a record hash, and $\sigma$ the integrity guarantee. Records and policy examples are illustrative.}
\label{fig:overview}
\end{figure*}

\section{Five Dimensions of Agent Auditability}
\label{sec:dimensions}

The central conceptual contribution of this paper is a five-dimensional
framework for agent auditability. Its purpose is not to add more terminology.
It is to answer a precise question: \emph{what, exactly, must be true of an
agent system before a post-deployment audit can produce a defensible verdict?}

A defensible audit yields a verdict of the form
\begin{equation}
\label{eq:verdict}
V = \bigl(\,
    \underbracket{s\vphantom{\mathrm{p}}}_{\scriptscriptstyle\text{Action Recov.}},\;
    \underbracket{c\vphantom{\mathrm{p}}}_{\scriptscriptstyle\text{Lifecycle Cov.}},\;
    \underbracket{\nu\vphantom{\mathrm{p}}}_{\scriptscriptstyle\text{Policy Check.}},\;
    \underbracket{r\vphantom{\mathrm{p}}}_{\scriptscriptstyle\text{Attribution}},\;
    \underbracket{\sigma\vphantom{\mathrm{p}}}_{\scriptscriptstyle\text{Integrity}}
    \,\bigr),
\end{equation}
where $s$ is the policy-relevant action under audit, $c$ is the execution
context in which that action occurred, $\nu \in \{\mathrm{comply},
\mathrm{violate}\}$ is the policy verdict under a stated policy $\pi$,
$r$ is the responsible component or responsibility chain, and
$\sigma$ is the integrity guarantee protecting the supporting record.
Each slot requires a distinct auditability condition, annotated beneath the
corresponding element in Eq.~\eqref{eq:verdict}. These five
conditions, i.e., Action Recoverability, Lifecycle Coverage, Policy Checkability,
Responsibility Attribution, and Evidence Integrity, are jointly necessary.
Removing any one renders the verdict either incomplete or untrustworthy.

Table~\ref{tab:dimensions} summarizes the framework. The five dimensions
can be read as the five questions an auditor must answer: \emph{what
happened}, \emph{in what context}, \emph{whether it complied}, \emph{who
was responsible}, and \emph{whether the evidence can be trusted}. Each
dimension is paired with two metrics: one for existence (does the evidence
appear at all?) and one for quality (is it strong enough to use?). The
subsections below develop each dimension through a concrete scenario and
define its metrics informally; the formal execution model and metric
definitions are in Appendix~\ref{app:metrics}.

\begin{table}[t]
  \caption{Five dimensions of agent auditability. Each dimension corresponds to
  a distinct slot in the audit verdict (Eq.~\ref{eq:verdict}) and a distinct
  auditor question. Accountability is not a sixth dimension; it is
  the derived property that emerges only when all five dimensions are
  sufficiently satisfied (\S\ref{subsec:synthesis}).}
  \label{tab:dimensions}
  \centering
  \small
  \begin{tabular}{@{}p{3.3cm}cp{3cm}p{5.2cm}@{}}
    \toprule
    \textbf{Dimension} & \textbf{Slot} & \textbf{Auditor question} & \textbf{Representative metrics} \\
    \midrule
    Action Recoverability
      & $s$
      & What happened?
      & \emph{ACR}: fraction of policy-relevant actions in the record. \newline
        \emph{RF}: fraction of required fields recoverable per recorded action. \\
    \addlinespace
    Lifecycle Coverage
      & $c$
      & In what context?
      & \emph{LPC}: fraction of observed lifecycle segments. \newline
        \emph{GB}: total unobserved lifecycle content. \\
    \addlinespace
    Policy Checkability
      & $\nu$
      & Compliance decidable?
      & \emph{SPDR}: fraction of policies decidable from the record. \newline
        \emph{ADL}: time from violation to determination. \\
    \addlinespace
    Responsibility Attribution
      & $r$
      & Who was responsible?
      & \emph{AC}: fraction of actions with full responsibility chain. \newline
        \emph{ACD}: average recovered chain length. \\
    \addlinespace
    Evidence Integrity
      & $\sigma$
      & Evidence trustworthy?
      & \emph{IS}: ordinal scale (none / append-only / hash-chained / signed). \newline
        \emph{VC}: time to verify integrity. \\
    \bottomrule
  \end{tabular}
\end{table}


\subsection{Action Recoverability}
\label{subsec:ar}

The first question in any audit is whether the system left a usable record
of the actions that matter. The relevant unit is not every latent model
state. It is the set of \emph{policy-relevant actions}: tool invocations,
external requests, file operations, database queries, approvals, and
delegation events. In practice, many existing agent systems and
observability tools record some actions but omit the fields needed to
reconstruct what actually happened (\S\ref{sec:alternatives}).

Consider an agent that calls a database API. The audit log records that a
call was made but omits the query arguments and the returned data. The
action is \emph{covered} but not \emph{recoverable}: a shallow log scores
well on coverage while scoring poorly on fidelity. We formalize this
distinction through two metrics: \emph{Action Coverage Rate (ACR)}, which
measures whether policy-relevant actions appear in the record at all, and
\emph{Record Fidelity (RF)}, which measures whether enough fields survive
for each recorded action to support reconstruction
(Appendix~\ref{app:ar-metrics}).

\subsection{Lifecycle Coverage}
\label{subsec:lc}

Recording individual actions is necessary but not sufficient. Even a fully
recorded action may be unauditable if the auditor cannot reconstruct the
execution context in which it occurred. Lifecycle Coverage operates over
execution \emph{phases} rather than individual steps and asks whether the
record covers the full execution structure. This is among the most
neglected dimensions in existing agent tools: retries, fallbacks,
approvals, and delegation handoffs are rarely represented as identifiable
phases in current traces (\S\ref{sec:alternatives}). The gap is structural
rather than incidental. A trace records what each step did, but not what it
relied on, so the dependency structure of a run has to be inferred rather than
read off the record~\citep{zhao2026grade}.

Return to the enterprise agent from \S\ref{sec:intro}. Suppose the agent
sends an email after querying customer records. The auditor can see both
actions. But the record omits that the agent first attempted to send a
different email, was blocked by a policy check, retried with modified
content, and received human approval on the second attempt. Without that
lifecycle context, the auditor cannot determine whether the final action was
a clean execution or a policy-circumventing retry.
\emph{Lifecycle Phase Coverage (LPC)} measures the fraction of execution
phases that are observed, and \emph{Gap Burden (GB)} measures how much
lifecycle structure is missing
(Appendix~\ref{app:lc-metrics}).

Note that LPC measures coverage of phases that actually occurred. It does
not by itself distinguish ``phase $p$ did not occur'' from ``phase $p$
occurred but was not recorded.'' Resolving that ambiguity may require
explicit phase-entry and phase-exit markers. We treat this as an open
measurement problem.

\subsection{Policy Checkability}
\label{subsec:pc}

A complete record of actions and lifecycle phases is still not auditable if
it cannot answer the policy question that motivated the audit. Policy
Checkability asks whether the record contains enough information for
compliance to be \emph{mechanically determined}. Existing runtime
enforcement tools can gate actions at execution time, but they do not
typically support post-hoc compliance verification from recorded evidence
(\S\ref{sec:alternatives}). The distinction matters: runtime blocking
prevents harm in real time, while policy checkability enables accountability
after the fact.

We focus on \emph{structural policies}: machine-checkable rules such as
``tool $X$ requires prior user approval'' or
``no external network call may follow access to data class $Y$ without
sanitization.'' A structural policy $\pi$ evaluated against the audit
record has three possible outcomes: $\mathrm{comply}$, $\mathrm{violate}$,
or $\bot$ (undecidable from the record). This third outcome, undecidable,
not merely unknown, is the reason the dimension exists. A policy is not
always either satisfied or violated. It can be \emph{impossible to decide}
because the record omits a required field.

\emph{Structural Policy Decidability Rate (SPDR)} measures the fraction of
policies that are decidable from the record, and \emph{Audit Detection
Latency (ADL)} measures the delay from violation to determination
(Appendix~\ref{app:pc-metrics}).

\begin{proposition}[Record schema determines policy decidability]
\label{prop:schema}
Let $\pi$ be a structural policy with required field set $F_\pi$,
and let $\mathcal{S}_\pi \subseteq \mathcal{S}$ be the set of execution
steps whose recorded fields could contribute to deciding $\pi$ (including,
e.g., approval steps for an approval-required policy).
Suppose $\mathcal{S}_\pi \neq \emptyset$.
If there exists $f \in F_\pi$ such that
$f \notin \widehat{F}(s_i;\mathcal{L})$
for all $s_i \in \mathcal{S}_\pi$,
then $\pi(\mathcal{L}) = \bot$.
\end{proposition}

This is not a deep result, but it is a consequential one: a single field
omitted from the record schema can render an entire class of policies
uncheckable, regardless of how many events are logged.
(Proof in Appendix~\ref{app:pc-metrics}.)

\subsection{Responsibility Attribution}
\label{subsec:ra}

Once the action is visible, the context is recovered, and the policy
verdict is determined, the remaining question is responsibility. In simple
systems, responsibility may appear local. In real deployments, it is often
a chain:
user $\rightarrow$ agent $\rightarrow$ skill $\rightarrow$ tool
$\rightarrow$ service, possibly with a human approval step in between.
While recent work on authenticated delegation addresses who is
\emph{permitted} to act, tracing who \emph{actually} caused a given outcome
across multi-agent delegation chains remains largely open
(\S\ref{sec:alternatives}).

Consider a multi-agent system where Agent~A delegates a task to Agent~B,
which invokes a third-party skill that calls an external API. The API
returns sensitive data. The audit log records Agent~B's API call, but not
that Agent~A initiated the task or that the skill was dynamically selected.
The immediate executor is visible, but the responsibility chain is broken.
\emph{Attribution Completeness (AC)} measures the fraction of actions with
a fully recovered chain, and \emph{Attribution Chain Depth (ACD)} measures
average recovered depth. When outcomes arise from joint behavior rather
than a single delegation sequence, attribution generalizes from chain
recovery to subgraph recovery over the interaction topology
(Appendix~\ref{app:ra-metrics}).

\subsection{Evidence Integrity}
\label{subsec:ei}

The previous four dimensions all depend on the audit record being
trustworthy. Yet integrity protection is nearly absent in existing agent
systems: among the approaches surveyed in \S\ref{sec:alternatives}, only
two provide any form of tamper-evident or cryptographically protected
records. Without integrity guarantees, a record that appears to satisfy
Action Recoverability, Lifecycle Coverage, Policy Checkability, and
Responsibility Attribution may have been silently modified after the fact.
Evidence Integrity is therefore foundational.

We define \emph{Integrity Strength (IS)} on an ordinal scale:

\begin{itemize}[leftmargin=*]
    \item \textbf{Level 0 (none):} no verification mechanism; entries can be modified without detection.
    \item \textbf{Level 1 (append-only):} entries cannot be deleted or reordered, but individual entries are not cryptographically bound.
    \item \textbf{Level 2 (hash-chained):} entries are linked by collision-resistant hashes, so modifying one entry invalidates subsequent links.
    \item \textbf{Level 3 (signed):} entries or batches are digitally signed, enabling third-party verification without relying on the original system.
\end{itemize}

We pair this with \emph{Verification Cost (VC)}: the time and compute
required to verify integrity over the full record. A mutable database table
with no append-only or cryptographic protection may appear operationally
convenient while still providing weak audit evidence
(Appendix~\ref{app:ei-metrics}).


\subsection{From Dimensions to Auditability}
\label{subsec:synthesis}

The five dimensions above are not an arbitrary list. They are derived from
the verdict structure in Eq.~\eqref{eq:verdict}: each slot requires exactly
one dimension, and no slot can be filled without its corresponding
condition. In \S\ref{sec:alternatives}, we confirm empirically that no
existing approach covers all five dimensions jointly, with Evidence
Integrity and Lifecycle Coverage as the most neglected. This subsection
shows why no fewer than five suffice, how the dimensions depend on each
other, and how they combine into a formal definition of auditability.

\paragraph{Why these five, and not six?}
Each dimension can fail independently while the others hold. A system can
log many events and still fail Policy Checkability because approval fields
or data-flow markers are absent. It can record individual actions and still
fail Lifecycle Coverage because retries, fallbacks, or escalations are not
represented as identifiable phases. It can support policy checks and still
fail Responsibility Attribution because delegated skill calls cannot be
linked back to their source. And it can satisfy the first four in
appearance while still failing Evidence Integrity if the record is mutable.

Possible candidates for a sixth dimension generally fall into one of two
categories. Some are already captured by the existing five. For example,
\emph{timeliness} is captured by Audit Detection Latency under Policy
Checkability, and \emph{interpretability} is largely captured by Record
Fidelity under Action Recoverability. Others are not dimensions of
auditability itself but constraints on audit design. For example,
\emph{privacy} constrains how evidence can be collected, retained, or
redacted, but it does not define an additional constituent of auditability.
Any property that might be proposed as a sixth dimension is therefore either
reducible to one of these five or is a design constraint rather than a
component of auditability itself.

\paragraph{Dependency structure.}
The five dimensions are not independent. Evidence Integrity is foundational:
without it, the other four cannot be trusted. Action Recoverability and
Lifecycle Coverage jointly enable Policy Checkability, which cannot operate
on actions or phases absent from the record. Responsibility Attribution is
parallel to Policy Checkability but independently necessary to complete the
verdict. The dependency is therefore: Evidence Integrity underpins the other
four; Action Recoverability and Lifecycle Coverage enable Policy
Checkability; Responsibility Attribution is separately required for
accountability.

\paragraph{Existence vs.\ diagnostic metrics.}
The auditability predicate thresholds only the existence-and-sufficiency
metrics: $\mathrm{ACR}$, $\mathrm{RF}$, $\mathrm{LPC}$, $\mathrm{GB}$,
$\mathrm{SPDR}$, $\mathrm{AC}$, and $\mathrm{IS}$. The remaining
metrics, $\mathrm{ADL}$, $\mathrm{ACD}$, and $\mathrm{VC}$, are
diagnostic. They characterize audit quality and operational burden after
auditability is established, but they do not determine if auditing is
possible in principle. A system can be auditable while still having high
detection latency, deep responsibility chains, or high verification cost.

\begin{definition}[Auditability]
\label{def:auditability}
Let $\Pi$ be a structural policy set and let
$\theta =
(\tau_{\mathrm{ACR}},\,
 \tau_{\mathrm{RF}},\,
 \tau_{\mathrm{LPC}},\,
 \tau_{\mathrm{GB}},\,
 \tau_{\mathrm{SPDR}},\,
 \tau_{\mathrm{AC}},\,
 \tau_{\mathrm{IS}})$
be a deployment-specific threshold vector.
Execution $X$ is \emph{auditable} with respect to $\Pi$,
record $\mathcal{L}$, and $\theta$ if all existence metrics meet or exceed
their thresholds (with $\mathrm{GB} \le \tau_{\mathrm{GB}}$).
The full formal statement with explicit inequalities is in
Appendix~\ref{app:def-auditability}.
\end{definition}

Note that the auditability predicate is monotone: if every thresholded
metric improves (or stays the same) and $\mathrm{GB}$ does not increase,
auditability is preserved.

Accountability becomes achievable when auditability holds and an auditor
can, from $\mathcal{L}$ alone, determine
$\pi_j(\mathcal{L}) \in \{\mathrm{comply},\mathrm{violate}\}$
for each relevant policy $\pi_j \in \Pi$, and recover the relevant
responsibility chain or interaction subgraph for each violation.
Accountability is therefore not a sixth dimension. It is the derived
property that becomes possible only when the five dimensions hold
simultaneously.

\section{Realizing the Five Dimensions}
\label{sec:lifecycle}

The five dimensions in \S\ref{sec:dimensions} define the conditions a
defensible audit must satisfy.
In practice, no single temporal vantage point can supply all of them.
Before deployment, one can inspect code and configuration but not realized
behavior.
During execution, one can observe and mediate live actions but only within
the runtime boundary.
After the fact, one can aggregate surviving evidence across systems but
cannot recreate evidence that was never captured or protected.
We therefore identify three classes of mechanism---\emph{detect},
\emph{enforce}, and \emph{recover}---each operating where its temporal
vantage point provides the strongest observational access and intervention
affordance.
\S\ref{sec:evidence} provides empirical support for each class in turn.

\begin{figure*}[!ht]
\centering
\begin{tikzpicture}[
  phase/.style={font=\footnotesize\scshape, text=black!50},
  arrow/.style={-{Stealth[length=5pt,width=4pt]}, thick, black!30},
  timelabel/.style={font=\scriptsize, text=black!45}
]

\node[rectangle, rounded corners=6pt, draw=detectblue!50, fill=detectblue!6,
      minimum width=3.8cm, text width=3.6cm, align=left, inner sep=7pt,
      line width=0.8pt] (detect) {
  {\color{detectblue!80!black}\small\bfseries Detect}\\[3pt]
  {\footnotesize\itshape\color{black!60} What could go wrong?}\\[5pt]
  {\footnotesize\color{black!75}\textbf{Sees:} code, configuration, skill metadata, permissions}\\[2pt]
  {\footnotesize\color{black!75}\textbf{Cannot:} observe realized runtime behavior}
};

\node[rectangle, rounded corners=6pt, draw=enforceteal!50, fill=enforceteal!6,
      minimum width=3.8cm, text width=3.6cm, align=left, inner sep=7pt,
      line width=0.8pt, right=0.4cm of detect] (enforce) {
  {\color{enforceteal!80!black}\small\bfseries Enforce}\\[3pt]
  {\footnotesize\itshape\color{black!60} Should this action proceed?}\\[5pt]
  {\footnotesize\color{black!75}\textbf{Sees:} live actions, data flows, policy context}\\[2pt]
  {\footnotesize\color{black!75}\textbf{Acts:} block, gate, approve, emit signed records}\\[2pt]
  {\footnotesize\color{black!75}\textbf{Cannot:} reach evidence after logs leave the system}
};

\node[rectangle, rounded corners=6pt, draw=recoverorange!50, fill=recoverorange!6,
      minimum width=3.8cm, text width=3.6cm, align=left, inner sep=7pt,
      line width=0.8pt, right=0.4cm of enforce] (recover) {
  {\color{recoverorange!80!black}\small\bfseries Recover}\\[3pt]
  {\footnotesize\itshape\color{black!60} What happened, and who is responsible?}\\[5pt]
  {\footnotesize\color{black!75}\textbf{Sees:} surviving records, cross-system traces}\\[2pt]
  {\footnotesize\color{black!75}\textbf{Cannot:} recreate evidence never captured or protected}
};

\node[phase, above=6pt of detect.north] {Pre-deployment};
\node[phase, above=6pt of enforce.north] {Runtime};
\node[phase, above=6pt of recover.north] {Post hoc};

\draw[arrow] (detect.east) -- (enforce.west);
\draw[arrow] (enforce.east) -- (recover.west);


\end{tikzpicture}
\caption{Information-and-intervention asymmetry across the three mechanism
classes. Each box details what the class can observe (\textbf{Sees}) and
what lies beyond its reach (\textbf{Cannot}), explaining why no single
temporal vantage point can satisfy all five auditability dimensions
(Table~\ref{tab:lifecycle-coverage}). Figure~\ref{fig:overview} introduces
the classes at a high level; this figure unpacks their operational
constraints.}
\label{fig:lifecycle}
\end{figure*}

\begin{table}[!ht]
    \caption{Which mechanism classes support which dimensions.
    \CIRCLE\ = direct support (evidence generation or post-hoc establishment),
    \LEFTcircle\ = partial or proxy signal,
    $\checkmark$ = verification target.
    No single column is fully filled: in practice, robust auditability
    draws on all three.}
  \label{tab:lifecycle-coverage}
  \centering
  \small
  \begin{tabular}{@{}lccc@{}}
    \toprule
    & \textbf{Detect} & \textbf{Enforce} & \textbf{Recover} \\
    \midrule
    Action Recoverability       & \LEFTcircle & \CIRCLE & \CIRCLE \\
    Lifecycle Coverage          & \LEFTcircle & \CIRCLE & \CIRCLE \\
    Policy Checkability         & \LEFTcircle & \CIRCLE & \CIRCLE \\
    Responsibility Attribution  & \LEFTcircle & \LEFTcircle & \CIRCLE \\
    Evidence Integrity & \LEFTcircle & \CIRCLE & \chmark \\
    \bottomrule
  \end{tabular}
\end{table}

Table~\ref{tab:lifecycle-coverage} shows that no single mechanism class
covers all five dimensions, and the reason is informational:
Detect can inspect code and configuration artifacts, but it cannot certify
properties of a realized execution $X$ or its actual record $\mathcal{L}$
(\S\ref{sec:dimensions}).
Enforce is the only temporal point that can both observe live actions and
create protected evidence as those actions occur---which is why it fills
four direct cells in the table.
Recover operates on the surviving record $\mathcal{L}$ as given; it cannot
add fields that were never captured or strengthen integrity guarantees
retroactively (Appendix~\ref{app:recovery-bounds}).
In practice, robust audit support therefore draws on all three, and the
evidence in \S\ref{sec:evidence} is organized accordingly: ecosystem
measurements validate detect, runtime feasibility validates enforce, and
missing-log recovery validates recover.

\subsection{Detect Before Deployment}

\paragraph{Mechanism: static analysis of code, configuration, and
supply-chain artifacts.}
Return to the enterprise agent from \S\ref{sec:intro}.
Before deployment, a static scan could flag that the email-sending skill has
no structured logging at the invocation site, that the customer-record query
exposes no approval hook, and that no signature or version pin protects the
skill's provenance metadata.
None of these findings prove that a violation will occur at runtime.
But each one identifies an auditability gap that, left unaddressed, will make
post-deployment accountability harder or impossible.

More generally, detect asks: Are policy-relevant actions instrumented?
Are approval paths visible? Are risky skills signed or version-pinned?
Table~\ref{tab:lifecycle-coverage} marks detect as partial (\LEFTcircle)
across all five dimensions because static artifacts can flag likely gaps but
cannot guarantee that those properties will hold at runtime.
The ecosystem scan in \S\ref{sec:evidence} provides quantitative evidence for
the scale of these gaps in the public agent ecosystem.

\subsection{Enforce During Execution}

\paragraph{Mechanism: runtime mediation of side-effecting actions.}
The \emph{enforce} class intercepts actions before they execute, evaluates
them against active policy, supports human approval where needed, and emits
structured, policy-relevant records.
In the enterprise-agent scenario, enforce is the mechanism that would
intercept the email-send action, check whether the drafted content violates a
data-handling policy, require human approval if configured, and---regardless
of the allow or block decision---emit a signed, timestamped record of the
action, the policy evaluation, and the outcome.

Enforcement does two things at once: it reduces risk in real time and
improves the quality of later auditing by generating complete, structured
evidence.
Runtime enforcement does not compete with auditability---it is one of the
main ways auditability becomes technically feasible.
Table~\ref{tab:lifecycle-coverage} marks enforce as direct (\CIRCLE) for
four dimensions.
It is only partial for Responsibility Attribution because runtime systems
often capture the immediate executor more reliably than the full upstream
delegation chain.
The runtime feasibility experiments in \S\ref{sec:evidence} provide
quantitative support for the practicality of this mechanism class.

\subsection{Recover After the Fact}

\paragraph{Mechanism: post-hoc reconstruction and responsibility assignment.}
The \emph{recover} class operates after deployment.
It reconstructs policy-relevant behavior from whatever evidence remains.
In the enterprise-agent scenario, suppose the email has already been sent and
the recipient reports an information disclosure the next day.
If enforce was active, the auditor queries the signed trace, reconstructs the
full execution path, and determines whether the data-handling policy was
violated.
But if the agent's output was forwarded into a ticketing system that stripped
execution metadata, or if the deployment spans vendors who each hold only a
partial trace, recovery must work from incomplete evidence.

This setting is not hypothetical.
In multi-party deployments, agent outputs are routinely copied into reports,
emails, or downstream systems where execution metadata is stripped.
No single party may hold the complete trace.
Recovery under missing or detached logs is therefore not an edge case but a
structural feature of realistic agent deployment.

Recover is the canonical setting in which accountability is established:
the auditor determines what happened, whether it complied with policy, and
who or what was responsible.
The missing-log recovery experiments in \S\ref{sec:evidence} show that
partial recovery of actions and responsibility is feasible even when conventional logs fail.

\paragraph{Why no single temporal point suffices.}
Auditability cannot be retrofitted at a single point in the system lifecycle.
A system that detects risks before deployment but generates no structured
evidence at runtime will fail on Policy Checkability and Evidence Integrity
when it matters most.
A system that enforces policy at runtime but cannot support post-hoc recovery
will fail when logs are incomplete, disputed, or detached.
The enterprise-agent scenario illustrates this directly: detect would have
flagged the missing hooks before deployment, enforce would have intercepted
the email and generated a signed record, and recover would have reconstructed
responsibility even after metadata was stripped.
No single class could have done all three.

\section{Evidence for the Auditability Gap}
\label{sec:evidence}

\begin{figure}[t]
  \centering
  \includegraphics[width=\textwidth]{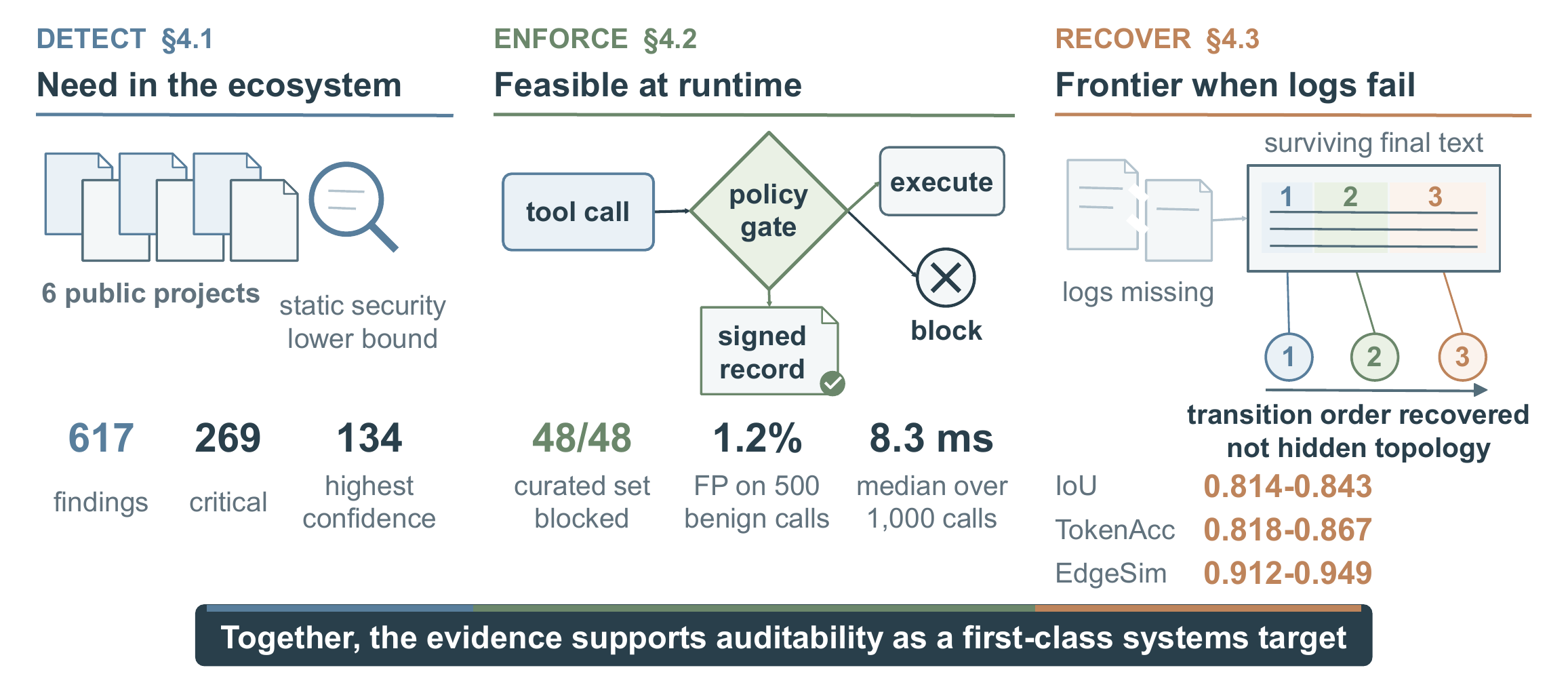}
  \caption{Three complementary evidence blocks support the position from
  different vantage points. Detect reports a conservative static-security
  lower bound from six public projects; Enforce summarizes a pre-execution
  policy gate and its curated evaluation; Recover reports IET ranges for
  segment and transition-order recovery when logs are unavailable. The blocks
  correspond to \S\ref{sec:lifecycle} and \S\ref{sec:evidence};
  Table~\ref{tab:evidence-dimensions} maps them to the five dimensions in
  Figure~\ref{fig:overview}. They are complementary studies, not one matched
  benchmark.}
  \label{fig:evidence-position}
\end{figure}

A position paper needs evidence, but not necessarily a single monolithic
benchmark. The right question is not whether one system already solves
auditing in every setting. It is whether there is enough evidence to make the
position difficult to dismiss. In this section, the evidence is intentionally
heterogeneous because it addresses three different objections, one for each
mechanism class identified in \S\ref{sec:lifecycle}. First, is the
auditability gap real in the public ecosystem (\emph{detect})? Second, are
auditable control points practical on the runtime execution path
(\emph{enforce})? Third, does accountability collapse when logs are
incomplete, redacted, or detached (\emph{recover})? We answer these
questions with three mutually reinforcing evidence blocks: an ecosystem lower
bound, a runtime feasibility layer, and a recovery frontier
(Figure~\ref{fig:evidence-position}).

\subsection{Ecosystem Lower Bound: the Public Default is Audit-blind}

\paragraph{Question.}
\textbf{Do publicly visible artifacts already show that minimal conditions for accountability are often absent?}

The first evidence block is intentionally conservative. We do not claim to
measure end-to-end auditability from public artifacts alone. Instead, we ask a
weaker and more defensible question: if basic \emph{security} gaps are already
visible in public agent code, can we at least conclude that the more demanding
requirements of post-deployment auditing are unlikely to be met? Security is a
prerequisite for auditability: a system with unvalidated tool inputs cannot
produce trustworthy action records, and a system without inter-agent
authentication cannot support responsibility attribution.

We draw on Agent Audit~\citep{zhang2026agentaudit}, a static security analysis
system whose paper describes 57 detection rules derived from 73 pattern types
and mapped to the OWASP Agentic Top~10 categories~(2026). Agent Audit performs tool-boundary taint tracking, MCP
configuration auditing, and semantic credential detection across Python agent
codebases. At ecosystem scale, the tool has been validated against
\SkillCount\ community-contributed skills on ClawHub.

For a detailed view, an archived public scan report produced with Agent Audit
v0.15.1~\citep{zhang2026agentaudit}
analyzed \ScanProjects\ prominent open-source agent
projects---OpenHands~\citep{wang2024openhands},
Generative Agents~\citep{park2023generative},
SWE-agent~\citep{yang2024sweagent},
Gorilla~\citep{patil2023gorilla},
MLAgentBench~\citep{huang2023mlagentbench},
and CodeAct~\citep{wang2024codeact}---representing the current
state of agentic AI development. Agent-audit identified \TotalFindings\
security findings, of which \CriticalFindings\ (\CriticalPct\%) were
classified as critical severity and \BlockTier\ findings were in the highest
confidence tier. The OWASP category breakdown reveals where the gaps concentrate:

\begin{figure}[t]
\centering
\begin{tikzpicture}[
  lbl/.style={font=\scriptsize, anchor=east, text=black!80},
  val/.style={font=\scriptsize\bfseries, anchor=west, text=black!70},
  dimtag/.style={font=\scriptsize, anchor=west}
]
\def\scale{0.0105}

\foreach \y/\cat/\cnt/\pct/\col/\dim in {
  0/{ASI-02 Tool Misuse}/397/64/dimAR/{Action Recov.},
  -0.58/{ASI-01 Goal Hijacking}/72/12/dimPC/{Policy Check.},
  -1.16/{ASI-07 Inter-Agent Comms}/66/11/dimEI/{Evidence Int.},
  -1.74/{ASI-05 Code Execution}/37/6/dimEI/{Evidence Int.},
  -2.32/{ASI-04 Credential Exposure}/27/4/dimRA/{Respons.\ Attr.},
  -2.90/{ASI-10 Rogue Agents}/12/2/dimLC/{Lifecycle Cov.},
  -3.48/{ASI-08 Cascading Failures}/4/{$<$1}/dimLC/{Lifecycle Cov.},
  -4.06/{ASI-09 Trust Exploitation}/2/{$<$1}/dimPC/{Policy Check.}%
}{
  \node[lbl] at (-0.15, \y) {\cat};
  \fill[\col, rounded corners=1.5pt]
    (0, \y-0.16) rectangle ({max(\cnt*\scale,0.08)}, \y+0.16);
  \node[val] at (\cnt*\scale+0.1, \y) {\cnt\enspace(\pct\%)};
  \node[dimtag, text=\col!80!black] at (\cnt*\scale+1.5, \y) {\dim};
}

\draw[black!25, line width=0.4pt] (-2.8, -4.50) -- (6.0, -4.50);
\node[font=\scriptsize\bfseries, anchor=east, text=black!80] at (-0.15, -4.76) {Total};
\node[font=\scriptsize\bfseries, anchor=west, text=black!70] at (0.1, -4.76) {\TotalFindings};

\draw[black!15, line width=0.4pt] (0, 0.35) -- (0, -4.35);

\end{tikzpicture}
\caption{Security findings from the archived Agent Audit v0.15.1 report across
\ScanProjects\ open-source agent projects by that release's legacy OWASP
Agentic category mapping (\TotalFindings\ total), color-coded by the
auditability dimension each category most directly undermines
(\S\ref{sec:dimensions}). Tool Misuse alone accounts for 64\% of all
findings.}
\label{fig:ecosystem-bars}
\label{tab:ecosystem-owasp} 
\end{figure}

Tool Misuse dominates: \ToolMisusePct\% of findings involve tool functions
that accept unvalidated input from the LLM, enabling injection, exfiltration,
and command execution. These are not exotic attack paths---they are the default
development pattern. General-purpose static analysis tools (Bandit, Semgrep)
achieve 0\% recall on MCP configuration vulnerabilities and substantially
lower recall on agent-specific patterns; agent-audit achieved
3--4$\times$ higher recall on a curated agent-vulnerability
benchmark~\citep{zhang2026agentaudit}.

These findings are security measurements, not direct auditability metrics. But
the connection is tight: every OWASP category maps to at least one auditability
dimension (Figure~\ref{fig:ecosystem-bars}), and systems that
fail these basic security checks are unlikely to produce the trustworthy
evidence that auditing requires. A supplementary platform-level scan of an AI
assistant with a skills marketplace found analogous patterns, including
supply-chain risks in community-contributed skill definitions
(Appendix~\ref{app:openclaw}). The public ecosystem's default is not merely
under-instrumented for auditing---it is under-secured for the actions agents
already take.

\subsection{Runtime Feasibility: Auditable Control Points are Practical}

\paragraph{Question.}
\textbf{Are auditable control points practical on the execution path?}

The second evidence block addresses a different objection: perhaps
auditability is conceptually attractive but too expensive, too invasive, or
too awkward to implement at runtime. Runtime feasibility evidence argues
otherwise.

We draw on the Aegis pre-execution firewall~\citep{yuan2026aegis}, which
inserts a framework-agnostic mediation point between the LLM's tool-call
decision and the underlying execution layer. Before any tool call executes,
Aegis recursively extracts string content from the call arguments, scans for
risk signals, and evaluates configurable policies. Each call receives one of
three decisions: \emph{allow}, \emph{block}, or \emph{pending} (escalated to
a human reviewer via a compliance dashboard).

Across 14 supported agent frameworks (Python, JavaScript, Go), Aegis
blocked all \AttackCases\ curated attack instances (spanning 7 OWASP
categories, including prompt injection, unauthorized tool use, and data
exfiltration) before side effects occurred. On \BenignCalls\ benign tool
calls (SELECT queries, file reads, API requests, text processing), it
yielded a \FalsePositiveRate\% false positive rate. End-to-end overhead
including SDK extraction, HTTP round-trip, classification, and policy
evaluation was \MedianLatency\ median across \LatencyTrials\ consecutive
interceptions (P95: 14.7\,ms, P99: 23.1\,ms---negligible relative to
typical LLM inference latency of 1--30\,s). The same runtime path
generated tamper-evident evidence via Ed25519-signed, SHA-256 hash-chained
records.

These aggregate numbers matter, but the dimension-level meaning matters more:
\begin{itemize}[leftmargin=*,itemsep=2pt]
  \item \emph{Action Recoverability}: the runtime layer records tool name, full arguments, output, timestamp, and policy decision for each intercepted call---supporting high record fidelity.
  \item \emph{Lifecycle Coverage}: allow, block, pending, and approval events are recorded as distinct execution states rather than collapsed into a single success-path trace.
  \item \emph{Policy Checkability}: structural policies are evaluated at interception time and the resulting decision is stored alongside the trace, making compliance mechanically decidable.
  \item \emph{Evidence Integrity}: Ed25519-signed, SHA-256 hash-chained records correspond to the strongest integrity level in our framework (Level~3).
  \item \emph{Responsibility Attribution}: only partially supported---current traces capture the immediate executor and session context but not the full upstream delegation chain.
\end{itemize}

These results do not show that one runtime layer solves agent safety.
That is not the claim.
They show something narrower and more important for this paper:
pre-execution control, structured evidence generation, approval workflows, and
tamper-evident logging are engineering-feasible with bounded overhead.
In other words, the mechanisms required for auditable agents are not
hypothetical.

\subsection{Recovery Frontier: Accountability under Missing Logs}

\paragraph{Question.}
\textbf{Does accountability collapse when logs are missing, redacted, or detached?}

The third evidence block addresses the hardest setting.
As argued in \S\ref{sec:lifecycle}, missing or detached logs are a structural
feature of realistic agent deployment, not an edge case.
If auditability depended only on ideal centralized logs, it would be too
brittle for multi-party deployment. Shared deployment strains the record in a
second way: when one agent serves several users through a shared state layer,
scope-bound artifacts from one user can be reapplied to another with no
attacker involved~\citep{yang2026crossuser}, which is exactly the kind of
cross-boundary reuse an auditor later has to untangle.

The recovery analysis uses implicit execution tracing (IET)~\citep{nian2026implicit}, a
method that embeds statistically controlled, key-conditioned signals into the
token distribution during generation. An auditor with the verification
registry can use these signals to recover which agent produced each segment of
a multi-agent output, even when explicit identity metadata, turn boundaries,
and orchestration logs have been removed.
At recovery time, a sliding-window scoring pass combined with change-point
detection reconstructs segment boundaries and the transition structure induced
by the linearized final output. It does not reconstruct the underlying agent
communication topology.

Topology is the right axis to vary here, since network structure governs how
information moves between agents and how far it travels~\citep{liu2026topology}.
Evaluated across chain, star, and tree coordination settings with 4--6 agents,
IET achieved:
\begin{itemize}[leftmargin=*,itemsep=2pt]
  \item \emph{Token attribution accuracy} of \RecoveryTokenAcc\ across evaluated settings.
  \item \emph{Segment overlap} (IoU) of \RecoveryIoU\ between recovered and ground-truth segments.
  \item \emph{EdgeSim} of \RecoveryEdgeSim\ for the transition structure induced by the linearized output.
\end{itemize}
IET was also evaluated under three degradation conditions: identity removal (all
agent identifiers stripped), boundary corruption (turn boundaries randomly
reshuffled), and redaction (sensitive information replaced with placeholders).
Under full identity removal, IET maintained non-trivial agent-level
attribution of 21.51--24.45\% across the three evaluated protocols, while
baselines reached 0--4.17\%.

The dimensional interpretation is deliberately narrow:
\begin{itemize}[leftmargin=*,itemsep=2pt]
  \item \emph{Responsibility Attribution} (\CIRCLE): segment attribution and transition-structure recovery let an auditor infer which component produced which portion of the output and the order in which attributed segments followed one another.
  \item \emph{Action Recoverability} (\LEFTcircle): segment overlap suggests that action-relevant boundaries can sometimes be reconstructed from surviving output alone.
  \item \emph{Policy Checkability} and \emph{Evidence Integrity} ({--}): not supported. Recovery cannot check policies or repair tampered records.
\end{itemize}
Recovery is therefore not a substitute for good logging.
It is a frontier capability that becomes relevant when the sixth question on
the Auditability Card---\emph{what happens when logs are missing, redacted, or
detached?}---has a non-trivial answer.

\begin{table}[t]
  \caption{Dimension-level coverage of the three evidence blocks.
  \CIRCLE\ = direct evidence,
  \LEFTcircle\ = partial or proxy evidence,
  {--} = not addressed.
  No single block covers all five dimensions; together, they cover
  the full framework from complementary directions.}
  \label{tab:evidence-dimensions}
  \centering
  \small
  \begin{tabular}{@{}lccc@{}}
    \toprule
    & \textbf{Ecosystem} & \textbf{Runtime} & \textbf{Recovery} \\
    & \textbf{(detect)} & \textbf{(enforce)} & \textbf{(recover)} \\
    \midrule
    Action Recoverability       & \LEFTcircle & \CIRCLE     & \LEFTcircle \\
    Lifecycle Coverage          & \LEFTcircle & \CIRCLE     & {--} \\
    Policy Checkability         & \LEFTcircle & \CIRCLE     & {--} \\
    Responsibility Attribution  & \LEFTcircle & \LEFTcircle & \CIRCLE \\
    Evidence Integrity          & \LEFTcircle & \CIRCLE     & {--} \\
    \bottomrule
  \end{tabular}
\end{table}

\paragraph{Synthesis.}
Table~\ref{tab:evidence-dimensions} summarizes the dimensional coverage of
the three evidence blocks.
The ecosystem lower bound provides proxy evidence across all five dimensions
but cannot certify any of them---it shows that public defaults are not
audit-ready.
The runtime feasibility block provides direct evidence for four dimensions,
confirming that auditable control points are practical.
The recovery frontier provides direct evidence for Responsibility Attribution
and partial evidence for Action Recoverability, showing that accountability
need not collapse completely when logging assumptions break down.
Together, the three blocks cover the full auditability framework from
complementary directions---enough to justify agent auditability as a
first-class research and systems target.

\section{Related Work and Alternative Views}
\label{sec:alternatives}

Table~\ref{tab:related} positions existing work against the five auditability
dimensions defined in \S\ref{sec:dimensions}. The table reveals two patterns.
First, no existing work covers all five dimensions jointly. Second, Evidence
Integrity and Lifecycle Coverage are the most neglected dimensions across all
categories. We discuss each category below and then engage the strongest
counterarguments directly.

\begin{table}[!ht]
  \caption{Positioning of existing work against the five auditability
  dimensions (\S\ref{sec:dimensions}).
  \CIRCLE\ = directly addressed,
  \LEFTcircle\ = partially addressed,
  {--} = not addressed.
  \textbf{Target}: what is audited.
  \textbf{Scope}: Pre = pre-deployment, Run = runtime, Post = post hoc, All = full lifecycle.
  Assessments reflect what each paper explicitly specifies or implements.
  Dimension abbreviations: AR = Action Recoverability, LC = Lifecycle Coverage,
  PC = Policy Checkability, RA = Responsibility Attribution,
  EI = Evidence Integrity (see Eq.~\ref{eq:verdict}).}
  \label{tab:related}
  \centering
  \small
  \setlength{\tabcolsep}{3.5pt}
  \begin{tabular}{@{}llllccccc@{}}
    \toprule
    \textbf{Category} & \textbf{Work} & \textbf{Target} & \textbf{Scope} & \textbf{AR} & \textbf{LC} & \textbf{PC} & \textbf{RA} & \textbf{EI} \\
    \midrule
    \multirow{3}{*}{\textit{Safety eval.}}
      & ToolEmu~\citep{ruan2024identifying}
        & Agent behavior & Pre
        & \LEFTcircle & {--} & \LEFTcircle & {--} & {--} \\
      & R-Judge~\citep{yuan2024rjudge}
        & Agent behavior & Pre
        & \LEFTcircle & {--} & \LEFTcircle & \LEFTcircle & {--} \\
      & Agent-SafetyBench~\citep{zhang2024agentsafetybench}
        & Agent behavior & Pre
        & \LEFTcircle & \LEFTcircle & \LEFTcircle & \LEFTcircle & {--} \\
    \addlinespace
    \multirow{2}{*}{\textit{Runtime enf.}}
      & AgentSpec~\citep{wang2025agentspec}
        & Agent actions & Run
        & \LEFTcircle & {--} & \LEFTcircle & {--} & {--} \\
      & AGrail~\citep{luo2025agrail}
        & Agent actions & Run
        & \LEFTcircle & {--} & \LEFTcircle & {--} & {--} \\
    \addlinespace
    \textit{Observability}
      & AgentOps~\citep{dong2024agentops}
        & Agent traces & Run
        & \LEFTcircle & \LEFTcircle & \LEFTcircle & {--} & {--} \\
    \addlinespace
    \multirow{2}{*}{\textit{Audit frmwk.}}
      & SMACTR~\citep{raji2020closing}
        & Dev.\ process & Pre
        & \LEFTcircle & \LEFTcircle & \LEFTcircle & \CIRCLE & {--} \\
      & Three-layer~\citep{mokander2023auditing}
        & LLM & All
        & \LEFTcircle & \LEFTcircle & \LEFTcircle & \LEFTcircle & {--} \\
    \addlinespace
    \multirow{3}{*}{\textit{Accountability}}
      & Auth.\ deleg.~\citep{south2025authenticated}
        & Agent deleg. & Run
        & \LEFTcircle & {--} & \CIRCLE & \CIRCLE & \LEFTcircle \\
      & Visibility~\citep{chan2024visibility}
        & Agent deploy. & Run
        & \LEFTcircle & \LEFTcircle & \LEFTcircle & \LEFTcircle & {--} \\
      & Audit trails~\citep{ojewale2026audit}
        & LLM lifecycle & All
        & \CIRCLE & \LEFTcircle & \LEFTcircle & \CIRCLE & \CIRCLE \\
    \midrule
    & \textbf{This paper}
        & \textbf{Agent exec.} & \textbf{All}
        & \CIRCLE & \CIRCLE & \CIRCLE & \CIRCLE & \CIRCLE \\
    \bottomrule
  \end{tabular}
\end{table}

\paragraph{Safety evaluation benchmarks.}
ToolEmu~\citep{ruan2024identifying} uses an LM-emulated sandbox to identify
risks across 36 high-stakes toolkits. R-Judge~\citep{yuan2024rjudge}
benchmarks safety risk awareness in multi-turn interactions, and
Agent-SafetyBench~\citep{zhang2024agentsafetybench} covers 16 agents across 8
risk categories. These three record structured agent traces (partial Action
Recoverability) and evaluate safety through LLM-based classifiers (partial
Policy Checkability). CatchBench instead compares failure detectability across
PRE, LIVE, and POST evidence~\citep{zhao2026catchbench}. None of the four
establishes all five dimensions, especially Evidence Integrity.

\paragraph{Runtime enforcement.}
AgentSpec~\citep{wang2025agentspec} provides a domain-specific language for
specifying runtime constraints with millisecond-level enforcement overhead.
AGrail~\citep{luo2025agrail} uses cooperative LLMs to iteratively refine
safety checks during test-time adaptation. Both gate actions at runtime
(partial Policy Checkability) and maintain in-memory trajectories (partial
Action Recoverability), but neither persists durable audit records, covers
the full execution lifecycle, traces responsibility chains, or protects
evidence integrity. As argued in \S\ref{sec:lifecycle},
enforcement is not an alternative to auditability---it is one of the main
ways auditability becomes technically feasible, because it generates
structured evidence as a byproduct.

\paragraph{Observability and audit frameworks.}
AgentOps~\citep{dong2024agentops} surveys 17 observability tools and
defines a span taxonomy for tracing agent behavior, but the paper itself
notes that ``trace links and interactions between different steps may have
been missed'' and provides no mechanisms for attribution or evidence
integrity. At the framework level,
SMACTR~\citep{raji2020closing} proposes an end-to-end internal audit
process with explicit stakeholder mapping (full Responsibility Attribution),
and~\citet{mokander2023auditing} distinguish governance, model, and
application audit layers. \citet{birhane2024auditing} document structural
barriers that can deny auditors the access and resources needed for meaningful
evaluation.
These frameworks target development processes, governance structures, and
model-level properties. They do not address the system-level properties
that arise when models are embedded in agent architectures that take
actions, delegate tasks, and interact with external services. Evidence
Integrity is unaddressed across all of them.

\paragraph{Agent accountability.}
The most closely related work addresses accountability at the agent level.
\citet{chan2024visibility} taxonomize what visibility measures are needed for
agent oversight and outline candidate mechanisms, but do not implement an
end-to-end audit system. WeClawArena records bounded runtime evidence for
cross-user collaboration and attack auditing~\citep{wang2026weclawarena}, but
does not specify all five properties. \citet{south2025authenticated}
propose agent-specific delegation tokens intended to support Policy
Checkability and Responsibility Attribution through machine-readable policies
and signed tokens, but cover only the authorization moment and not the full
execution lifecycle.
At the audit-record level, \citet{ojewale2026audit} use
SHA-256 hash-chained, append-only records with explicit actor fields,
covering five model lifecycle stages with strong Evidence Integrity.
However, their lifecycle is the \emph{model} lifecycle (pretraining,
fine-tuning, deployment, monitoring), not the \emph{agent runtime}
lifecycle (tool calls, sub-agent delegation, retries, approval workflows,
fallback paths). Their event taxonomy includes
\texttt{FineTuneStart} and \texttt{DeploymentCompleted} but not
the side-effecting actions, dynamic skill invocations, and multi-step
execution chains that define agent behavior.
Their Policy Checkability is also partial: governance checkpoints are
recorded as first-class events, but no policy engine or formal verification
mechanism is specified.
Our framework is complementary: it defines auditability conditions
specifically for the agent runtime setting, with formal metrics that make
each dimension measurable across the detect--enforce--recover lifecycle.

\paragraph{Software audit infrastructure and documentation.}
Tamper-evident logging~\citep{crosby2009efficient} provides the
cryptographic foundation for our Evidence Integrity dimension.
Sigstore~\citep{newman2022sigstore} demonstrates that verifiable signing can
be made practical at ecosystem scale, a model transferable to agent action
attestation. Software supply-chain security~\citep{ohm2020backstabbers}
faces similar provenance challenges, compounded in agent ecosystems by
dynamic skill selection. Model cards~\citep{mitchell2019model} and
datasheets~\citep{gebru2021datasheets} established the norm that ML
artifacts should ship with structured documentation. Our minimal
Auditability Card (\S\ref{sec:agenda}) extends this paradigm from static
artifacts to deployed agent executions.


\medskip

We now engage four alternative views that a skeptical reviewer might
reasonably hold.

\paragraph{Alternative view 1: observability is enough.}
Modern tracing and dashboarding systems provide useful infrastructure.
AgentOps~\citep{dong2024agentops} defines detailed span types and
metadata for agent monitoring. But as Table~\ref{tab:related} shows,
observability tools leave Responsibility Attribution and Evidence Integrity
entirely unaddressed. A dashboard that records prompts and tool names but
omits approval events, data categories, caller identities, or integrity
guarantees can still fail Auditability on multiple dimensions. Observability
tells an operator \emph{that} something happened. Auditability asks whether
the record is sufficient to determine \emph{what} happened, \emph{whether}
it complied with policy, and \emph{who} was responsible.

\paragraph{Alternative view 2: runtime blocking matters more than post hoc
audit.}
Runtime blocking matters a great deal. Systems like
AgentSpec~\citep{wang2025agentspec} and AGrail~\citep{luo2025agrail}
demonstrate that enforcement is practical and effective. But blocking does
not remove the need for auditing. A blocked action still needs an
explanation. An allowed action still needs a trace. A human override still
needs provenance. Blocking reduces risk in real time. Auditing makes behavior
answerable after the fact. Our runtime feasibility evidence
(\S\ref{sec:evidence}) confirms this: pre-execution mediation adds
single-digit-millisecond overhead while generating tamper-evident records
as a byproduct.

\paragraph{Alternative view 3: stronger alignment will make auditability
less important.}
This view assumes that harmful behavior is mainly a model-internal problem.
In deployed agent systems, many failures arise from dynamic tool behavior,
changing external services, skill updates, human approvals, and multi-agent
composition~\citep{ruan2024identifying,zhang2024agentsafetybench}. These
are deployment properties, not model properties. Refusal training, the
standard alignment lever, also transfers poorly to this setting: agentic harm
lies in the gap between the authority an action exercises and the authority
the user granted, which is absent from the text the model
sees~\citep{li2026actionalignment}. Defense-trained models can lose agent
competence without gaining protection against capable
attacks~\citep{li2026autonomytax}. Better alignment helps,
but it does not by itself establish accountability for what a specific
deployed system actually did. Even a perfectly aligned model embedded in a
poorly instrumented agent system remains unauditable.

\paragraph{Alternative view 4: auditability is too costly or too invasive.}
Poor evidence design can indeed be expensive or privacy-invasive. But the
correct response is not to abandon auditability. It is to design evidence
better. Selective recording, policy-scoped schemas, redaction, encryption,
hashed summaries~\citep{crosby2009efficient}, and metadata-light recovery
are all ways to reduce cost while preserving accountability. Our own
evidence (\S\ref{sec:evidence}) shows that pre-execution mediation adds
\MedianLatency\ median overhead, and that partial evidence about actions and
responsibility remains recoverable even when conventional logs are
missing~\citep{nian2026implicit}.
The software supply-chain community has shown that verifiable signing can be
made practical at scale~\citep{newman2022sigstore}. Cost and privacy are
evidence-design constraints, not reasons to give up on answerable systems.

\section{Auditability Card and Open Problems}
\label{sec:agenda}

The evidence in the preceding sections shows that the auditability gap is
real, that closing it is engineering-feasible, and that partial recovery of
actions and responsibility
survives even when logs fail. The remaining question is adoption. This
section proposes two deliverables: an \emph{Auditability Card} that agent
systems can report immediately, and six open research problems that the
community must solve to make auditable agents the default.

%
%

\paragraph{The Auditability Card.}
Table~\ref{tab:card} defines the Auditability Card: six questions that any
agent paper, benchmark, framework, or skill ecosystem should answer.
Q1--Q5 correspond to the five auditability dimensions
(\S\ref{sec:dimensions}); Q6 stress-tests what happens when logging
assumptions break down. The rightmost column shows an illustrative partial
card for a runtime firewall~\citep{yuan2026aegis}, demonstrating the format
on a real system.

\begin{table}[t]
  \caption{The Auditability Card. Q1--Q5 map to the five dimensions; Q6 is a
  stress test. The rightmost column is an illustrative partial card for
  Aegis~\citep{yuan2026aegis}, not a canonical answer.}
  \label{tab:card}
  \centering
  \small
  \setlength{\tabcolsep}{3pt}
  \begin{tabular}{@{}p{2.3cm}p{3.8cm}p{5.6cm}@{}}
    \toprule
    \textbf{Question} & \textbf{What to disclose} & \textbf{Illustrative example (Aegis)} \\
    \midrule
    Q1: Actions \newline {\scriptsize (Action Recov.)}
      & What policy-relevant actions does the system record?
      & Tool calls with name, full arguments, output, timestamp, and policy decision \\
    \addlinespace
    Q2: Phases \newline {\scriptsize (Lifecycle Cov.)}
      & Which execution phases are covered?
      & Allow, block, pending, and approval as distinct states; delegation chains not covered \\
    \addlinespace
    Q3: Policies \newline {\scriptsize (Policy Check.)}
      & What policies can be mechanically checked?
      & Configurable structural rules; decision stored alongside each call \\
    \addlinespace
    Q4: Attribution \newline {\scriptsize (Respons.\ Attr.)}
      & What responsibility chain is available?
      & Immediate executor and session context; upstream delegation chain partial \\
    \addlinespace
    Q5: Integrity \newline {\scriptsize (Evidence Int.)}
      & What protects the record from modification?
      & Level~3: Ed25519-signed, SHA-256 hash-chained records \\
    \addlinespace
    Q6: Missing logs
      & What happens when logs are missing or detached?
      & No built-in recovery; depends on external evidence \\
    \bottomrule
  \end{tabular}
\end{table}

The card is deliberately compact. Its value lies not in comprehensiveness but
in forcing disclosure: a system can answer these questions well or badly, but
it should not be allowed to answer them ambiguously.

We envision four adoption paths:
\begin{itemize}[leftmargin=*,itemsep=2pt]
  \item \textbf{Papers} claiming safety, reliability, or deployment readiness
    should include the card in their evaluation or appendix.
  \item \textbf{Benchmarks} should require card-level disclosure alongside
    task-performance metrics.
  \item \textbf{Frameworks} should auto-generate a partial card from runtime
    configuration, specifying which phases are logged and what integrity level
    is provided.
  \item \textbf{Skill ecosystems} should require card-level provenance
    metadata as a publication
    prerequisite~\citep{ohm2020backstabbers}.
\end{itemize}

%
%

\paragraph{Open problems.}
The card addresses reporting. The harder question is what the community must
still solve. We identify six open research problems, organized by the
mechanism classes from \S\ref{sec:lifecycle}. Each is grounded in a specific
evidence gap from this paper.

\paragraph{Detect.}
\emph{OP1: Predicting auditability gaps from code.}
Our ecosystem scan (\S\ref{sec:evidence}) uses security findings as proxies
for auditability gaps. Can static analysis directly predict which of the five
dimensions will be under-supported at runtime, before deployment, by
analyzing code structure, logging instrumentation, and event-schema coverage?

\emph{OP2: Minimal provenance for dynamic skills.}
Supply-chain risks appeared across all scanned projects, and the skill layer is
itself a privilege boundary that current models routinely
exceed~\citep{li2026fortis}. Large skill libraries have begun to treat source
grounding as an explicit traceability guarantee, mapping each retained claim
back to the source it came from~\citep{sha2026skillcenter}, which suggests such
a record is constructible at library scale. What is the minimal
provenance record---signature, version, permission scope, auditability
card, that a dynamically selected skill must carry to support post-hoc
responsibility attribution?

\paragraph{Enforce.}
\emph{OP3: Full-chain attribution at runtime.}
Our runtime evidence shows that Responsibility Attribution is only partially
supported: the immediate executor is captured but the upstream delegation
chain is not (\S\ref{sec:evidence}). How can a mediation layer capture the
full responsibility chain across multi-agent delegation without requiring all
frameworks to share a single event schema?

\emph{OP4: Semantic policy decidability.}
Our framework restricts Policy Checkability to structural policies
(\S\ref{subsec:pc}). What classes of semantic policies, e.g., ``the agent
should not disclose information that could identify the customer'', can be
made mechanically decidable from the audit record, and at what cost in record
schema complexity?

\paragraph{Recover.}
\emph{OP5: Adversarial recovery.}
Our recovery evidence tests benign degradation: identity removal, boundary
corruption, and redaction (\S\ref{sec:evidence}). How robust is
metadata-light recovery when the degradation is adversarial, when an
attacker deliberately targets the attribution signal? What are the
information-theoretic limits of recovery without explicit logs?

\emph{OP6: Cross-party audit aggregation.}
Multi-party deployments structurally fragment evidence
(\S\ref{sec:lifecycle}). How can multiple parties, each holding a partial
trace with independent integrity guarantees, produce a joint audit verdict
without any single party holding the complete record?

\paragraph{From position to practice.}
The Auditability Card provides a reporting standard that can be adopted
immediately. The six open problems define a research agenda for making
auditable agents the default rather than the exception.

\section{Limitations}
\label{sec:limitations}

\paragraph{Evidence drawn from the authors' own tools.}
All three evidence blocks rely on tools developed by the authors:
agent-audit~\citep{zhang2026agentaudit} for the ecosystem scan,
Aegis~\citep{yuan2026aegis} for runtime feasibility, and
IET~\citep{nian2026implicit} for recovery experiments. While the tools
are open-source and the experiments are reproducible, the position
would be strengthened by independent replication or by evidence from
tools developed outside this group. In addition, the ecosystem scan
provides lower-bound proxy evidence from security findings rather than
direct measurement of end-to-end auditability. All tools and data
used in this paper are or will be publicly available to facilitate
independent replication.

\paragraph{No end-to-end audit.}
The three evidence blocks validate individual mechanism classes in
isolation. We have not demonstrated a complete audit workflow in which
all five dimensions are measured on a single deployed system and a
defensible verdict is produced end to end. Such a demonstration would
require a deployment with ground-truth violations, a full-stack
evidence pipeline, and an evaluator---a significant engineering and
experimental undertaking that we leave to future work.

\paragraph{Scale and diversity of evidence.}
The ecosystem scan covers six open-source projects; the runtime
evaluation uses 48 curated attacks and 500 benign calls; the recovery
experiments test 4--6 agents in controlled topologies. These are
sufficient to support the paper's claims as lower bounds and feasibility
demonstrations, but they do not constitute a comprehensive benchmark
across the diversity of real-world agent architectures, programming
languages, or orchestration patterns.

\paragraph{Open-source systems only.}
All evidence is drawn from open-source agent projects. Commercial and
enterprise agent deployments, where auditability arguably matters
most, are not examined. Proprietary systems may have internal audit
infrastructure not visible in public code, or may face additional
constraints (vendor lock-in, cross-organizational trust boundaries)
that our framework does not yet address empirically.

\paragraph{Threshold calibration.}
The auditability predicate (Definition~\ref{def:auditability}) depends
on a deployment-specific threshold vector $\theta$, but the paper
provides no empirical guidance on how to calibrate these thresholds.
What ACR or SPDR value is ``good enough'' likely depends on the risk
profile, regulatory context, and policy set of a specific deployment.
Developing principled calibration methods, whether through
domain-expert elicitation, regulatory mapping, or empirical
benchmarking, remains open.

\paragraph{Structural policies only.}
The current formalization and all empirical support are limited to
structural, machine-checkable policies. Many real-world compliance
questions are semantically rich or context-dependent, e.g., ``the agent
should not disclose information that could identify the customer'', and
remain outside what the current record schema can decide.
Open problem OP4 (\S\ref{sec:agenda}) identifies a research path for
relaxing this boundary, but the limitation is present in all results
reported here.

\paragraph{Completeness of the five dimensions.}
The argument that five dimensions are necessary and sufficient is
grounded in the verdict structure (Eq.~\ref{eq:verdict}) and an
informal reducibility argument (\S\ref{subsec:synthesis}). We do not
provide a formal completeness proof. It is possible that future agent
architectures, for example, embodied agents with physical
side effects or agents operating under real-time safety
constraints, may surface dimensions not reducible to the current five.

\paragraph{Privacy and data-minimization tension.}
Comprehensive audit records can conflict with data-minimization
principles such as those in GDPR and similar regulations. The paper
acknowledges privacy as a design constraint
(\S\ref{sec:alternatives}), but does not develop concrete mechanisms
for reconciling high-fidelity audit records with data-minimization
requirements. Selective recording, access-controlled disclosure,
redaction policies, and cryptographic commitments that preserve
integrity while limiting exposure are plausible directions, but this
paper does not establish which privacy-preserving mechanisms can retain
audit-grade fidelity.

\section{Conclusion}
\label{sec:conclusion}

This paper argues that once an agent system can act in the world,
\emph{auditability}---the ability to reconstruct what it did, check whether
it complied with policy, and attribute responsibility---should be treated as
a first-class design requirement, not an afterthought.

We made this position concrete. Five dimensions define what a defensible
post-deployment audit requires. Three mechanism classes (detect, enforce,
recover) show why no single temporal vantage point can supply all five.
Layered evidence from ecosystem scans, runtime mediation, and missing-log
recovery supports the claim that the auditability gap is real, that core
auditability mechanisms are engineering-feasible, and that partial
accountability can survive even when conventional logs fail. The Auditability Card and six open problems in
this paper offer a path from position to practice.

We believe the implications extend beyond the specific framework proposed
here. If the community adopts auditability as a standard evaluation
criterion, alongside accuracy, safety, and efficiency, it will reshape how
agent systems are designed, documented, and deployed. Frameworks will need to
emit structured, integrity-protected evidence by default. Benchmarks will
need to measure not only whether agents succeed at tasks, but whether their
behavior remains reconstructable afterward. Skill ecosystems will need
provenance metadata as a publishing prerequisite.

The field has invested heavily in making agents capable and safe. The
complementary question, i.e., whether their actions remain answerable, is now
urgent. Auditability is not a tax on agent development. It is a foundation for trust, accountability, and responsible
deployment.

\clearpage
\newpage


\bibliographystyle{plainnat}
\bibliography{references,yue-zhao,working}
\clearpage
\newpage

\appendix
\setcounter{figure}{0}
\setcounter{table}{0}
\setcounter{footnote}{0}                                                                                               
\renewcommand{\thefigure}{\Alph{section}\arabic{figure}}
\renewcommand{\thetable}{\Alph{section}\arabic{table}}

\section*{Supplementary Material for Auditable Agents}

\section{Appendix Overview}
\label{app:overview}

This appendix provides the formal machinery underlying the five auditability
dimensions (\S\ref{app:metrics}), recovery bounds
(\S\ref{app:recovery-bounds}), and a supplementary platform-level security
scan (\S\ref{app:openclaw}). Full evidence protocols are documented in the
respective tool papers: agent-audit~\citep{zhang2026agentaudit} for the
ecosystem scan, Aegis~\citep{yuan2026aegis} for runtime feasibility, and
IET~\citep{nian2026implicit} for recovery experiments.

\section{Formal Execution Model and Metric Definitions}
\label{app:metrics}

This section provides the formal machinery underlying the five auditability
dimensions introduced in \S\ref{sec:dimensions}. The main text defines each
dimension through intuition and concrete examples; here we give precise
mathematical definitions.

\subsection{Shared execution--record model}
\label{app:exec-model}

An \emph{agent execution} is a tuple
\[
X = (\mathcal{A}, \mathcal{S}, \psi, \rho),
\]
where $\mathcal{A} = \{a_1,\ldots,a_m\}$ is the set of participating
components (agents, tools, skills, services, or human principals), and
$\mathcal{S} = (s_1,\ldots,s_n)$ is the sequence of execution steps.
Each step is
\[
s_i = (\mathrm{type}_i,\ \mathrm{in}_i,\ \mathrm{out}_i,\ t_i,\ \mathrm{ctx}_i),
\]
where $\mathrm{type}_i$ is an action type, $\mathrm{in}_i$ and
$\mathrm{out}_i$ are the step input and output, $t_i$ is a timestamp, and
$\mathrm{ctx}_i$ is execution context, such as approval state, caller chain,
or phase label.
The function $\psi(s_i)$ assigns each step to a lifecycle phase, and
$\rho(s_i)$ assigns each step a responsibility chain from the immediate
executor back to the originating principal.

An \emph{audit record} for execution $X$ is a tuple
\[
\mathcal{L} = (\mathcal{R}, \sigma),
\]
where $\mathcal{R} = (r_1,\ldots,r_k)$ is a sequence of record entries and
$\sigma$ is an integrity mechanism, which may be null.
Each record entry $r_j$ is a partial observation of one or more execution
steps.
We write $\phi(j) \subseteq \mathcal{S}$ for the set of execution steps
observed by entry $r_j$, and $\mathrm{fields}(r_j)$ for the set of fields
preserved by that entry.
For any step $s_i$, let
\[
\widehat{F}(s_i;\mathcal{L}) =
\bigcup_{j:\, s_i \in \phi(j)} \mathrm{fields}(r_j)
\]
denote the set of fields about $s_i$ that are recoverable from the audit
record.

\subsection{Action Recoverability metrics}
\label{app:ar-metrics}

Let $\mathcal{S}_{\mathrm{rel}} \subseteq \mathcal{S}$ denote the
policy-relevant steps in execution $X$.
This set is specified by the deployment context and is defined independently
of any particular policy set $\Pi$.
We define
\[
\mathrm{ACR}(X,\mathcal{L}) =
\frac{|\{s_i \in \mathcal{S}_{\mathrm{rel}} :
\widehat{F}(s_i;\mathcal{L}) \neq \emptyset\}|}
{|\mathcal{S}_{\mathrm{rel}}|}.
\]

For each step $s_i$, let $F_{\mathrm{obs}}(s_i)$ be the minimal set of
observational fields needed to reconstruct its externally meaningful effect.
We then define
\[
\mathrm{RF}(X,\mathcal{L}) =
\frac{1}{|\mathcal{S}_{\mathrm{cov}}|}
\sum_{s_i \in \mathcal{S}_{\mathrm{cov}}}
\frac{|F_{\mathrm{obs}}(s_i) \cap \widehat{F}(s_i;\mathcal{L})|}
{|F_{\mathrm{obs}}(s_i)|},
\]
where
$\mathcal{S}_{\mathrm{cov}} =
\{s_i \in \mathcal{S}_{\mathrm{rel}} :
\widehat{F}(s_i;\mathcal{L}) \neq \emptyset\}$.

\paragraph{Edge cases.}
If $|\mathcal{S}_{\mathrm{rel}}| = 0$ (no policy-relevant actions occurred),
we define $\mathrm{ACR} = 1$ by convention: there is nothing to record and
nothing is missing.
If $|\mathcal{S}_{\mathrm{cov}}| = 0$ and $|\mathcal{S}_{\mathrm{rel}}| > 0$
(policy-relevant actions exist but none are covered),
we define $\mathrm{RF} = 0$.
If $|\mathcal{S}_{\mathrm{rel}}| = 0$, both ACR and RF are~1 by convention:
there is nothing to record and nothing to recover.

\paragraph{Why both metrics are needed.}
ACR and RF are not redundant. A system can achieve $\mathrm{ACR} = 1$
(every policy-relevant action appears in the record) while
$\mathrm{RF} \approx 0$ (each entry records only that a tool was called,
omitting arguments, output, caller identity, and approval context).
Such a system \emph{covers} every action but \emph{recovers} none.
Conversely, a system with $\mathrm{ACR} \ll 1$ but high RF for its
covered actions records fewer events but records them well.
The two metrics capture orthogonal failure modes: missing events versus
missing fields.

\subsection{Lifecycle Coverage metrics}
\label{app:lc-metrics}

Let $\mathcal{I}(X) = \{I_1,\ldots,I_q\}$ be the maximal contiguous lifecycle
segments induced by $\psi$, where each segment $I_\ell$ contains consecutive
steps with the same phase label.
We say that a segment $I_\ell$ is \emph{observed} if the audit record contains
enough information to infer both its existence and its phase label.
We define
\[
\mathrm{LPC}(X,\mathcal{L}) =
\frac{|\{I_\ell \in \mathcal{I}(X) : I_\ell \text{ is observed}\}|}
{|\mathcal{I}(X)|},
\]
and
\[
\mathrm{GB}(X,\mathcal{L}) =
\sum_{I_\ell \in \mathcal{I}(X):\, I_\ell \text{ not observed}} |I_\ell|,
\]
where $|I_\ell|$ denotes either duration or step count.
A deployment must fix one unit (duration or step count) and use the same
unit in the $\mathrm{GB}$ threshold $\tau_{\mathrm{GB}}$.

\paragraph{Edge case.}
If $|\mathcal{I}(X)| = 0$ (the execution has no identifiable lifecycle
segments), we define $\mathrm{LPC} = 1$ and $\mathrm{GB} = 0$: there is no
lifecycle structure to miss.

\paragraph{Why both metrics are needed.}
LPC and GB measure different aspects of lifecycle coverage. A system with
$\mathrm{LPC} = 0.9$ might be missing a single short phase
($\mathrm{GB}$ = 1 step) or a single very long phase
($\mathrm{GB}$ = 1{,}000 steps). The coverage rate is the same, but the
gap burden differs by three orders of magnitude. LPC captures
\emph{how many} phases are observed; GB captures \emph{how much} execution
content is missing.

\subsection{Policy Checkability metrics}
\label{app:pc-metrics}

Let $\Pi = \{\pi_1,\ldots,\pi_k\}$ be a policy set, where each
$\pi_j(\mathcal{L}) \in
\{\mathrm{comply},\ \mathrm{violate},\ \bot\}$.
We define
\[
\mathrm{SPDR}(\Pi,\mathcal{L}) =
\frac{|\{\pi_j \in \Pi : \pi_j(\mathcal{L}) \neq \bot\}|}
{|\Pi|}.
\]

For each violated and detected policy $\pi_j$, let
$t_{\mathrm{violate}}(\pi_j)$ denote the timestamp of the earliest
action that violates $\pi_j$, and let $t_{\mathrm{detect}}(\pi_j,
\mathcal{L})$ denote the earliest time at which the violation is
determinable from the audit record. We define
\[
\mathrm{ADL}(\pi_j,\mathcal{L}) =
t_{\mathrm{detect}}(\pi_j,\mathcal{L}) - t_{\mathrm{violate}}(\pi_j).
\]

\paragraph{Edge case.}
If $|\Pi| = 0$ (no structural policies are specified), we define
$\mathrm{SPDR} = 1$: there are no policies to check and none are
undecidable. ADL is undefined when no violation is detected.

\paragraph{Proof of Proposition~\ref{prop:schema}.}
By definition, $\pi$ is decidable from $\mathcal{L}$ if
$\pi(\mathcal{L}) \in \{\mathrm{comply}, \mathrm{violate}\}$.
Let $\mathcal{S}_\pi$ be the evidence steps for $\pi$ as defined in the
proposition. Decidability requires that the union of recovered fields
across all evidence steps includes every required field:
$F_\pi \subseteq \bigcup_{s_i \in \mathcal{S}_\pi}
\widehat{F}(s_i;\mathcal{L})$.
If there exists $f \in F_\pi$ such that $f \notin
\widehat{F}(s_i;\mathcal{L})$ for every $s_i \in \mathcal{S}_\pi$, then
$f \notin \bigcup_{s_i \in \mathcal{S}_\pi}
\widehat{F}(s_i;\mathcal{L})$, the inclusion fails, and
$\pi(\mathcal{L}) = \bot$. \hfill$\square$

\subsection{Responsibility Attribution metrics}
\label{app:ra-metrics}

For each step $s_i$, let
$\rho(s_i) =
(a_i^{(1)}, a_i^{(2)}, \ldots, a_i^{(d_i)})$
denote the ground-truth responsibility chain, ordered from the immediate
executor back to the originating principal.
Let $\widehat{\rho}(s_i;\mathcal{L})$ denote the longest recoverable prefix.
We define
\[
\mathrm{AC}(X,\mathcal{L}) =
\frac{|\{s_i \in \mathcal{S}_{\mathrm{rel}} :
|\widehat{\rho}(s_i;\mathcal{L})| = |\rho(s_i)|\}|}
{|\mathcal{S}_{\mathrm{rel}}|},
\]
and
\[
\mathrm{ACD}(X,\mathcal{L}) =
\frac{1}{|\mathcal{S}_{\mathrm{rel}}|}
\sum_{s_i \in \mathcal{S}_{\mathrm{rel}}}
|\widehat{\rho}(s_i;\mathcal{L})|.
\]

\paragraph{Edge case.}
AC and ACD share the denominator $|\mathcal{S}_{\mathrm{rel}}|$ with ACR.
If $|\mathcal{S}_{\mathrm{rel}}| = 0$, we define $\mathrm{AC} = 1$ and
$\mathrm{ACD} = 0$ by the same convention: there are no actions to
attribute.

When outcomes arise from joint behavior of multiple components rather than a
single delegation sequence, responsibility is no longer a chain but a subgraph
of the interaction topology
$G = (\mathcal{A}, E)$,
where $(a_u, a_v) \in E$ indicates that component $a_u$'s output influenced
component $a_v$'s action.
In such settings, Attribution Completeness generalizes from chain recovery to
subgraph recovery.

\subsection{Evidence Integrity metrics}
\label{app:ei-metrics}

Integrity Strength (IS) is defined on the ordinal scale in \S\ref{sec:dimensions}.
Verification Cost is
\[
\mathrm{VC}(\mathcal{L}) =
\text{wall-clock time required to verify } \sigma
\text{ over the full record.}
\]

\subsection{Formal definition of auditability}
\label{app:def-auditability}

\begin{definition}[Auditability --- full formal statement]
Let $\Pi$ be a structural policy set and let
\[
\theta =
(\tau_{\mathrm{ACR}},\,
 \tau_{\mathrm{RF}},\,
 \tau_{\mathrm{LPC}},\,
 \tau_{\mathrm{GB}},\,
 \tau_{\mathrm{SPDR}},\,
 \tau_{\mathrm{AC}},\,
 \tau_{\mathrm{IS}})
\]
be a deployment-specific threshold vector.
Execution $X$ is \emph{auditable} with respect to policy set $\Pi$,
record $\mathcal{L}$, and threshold vector $\theta$ if
\begin{align*}
\mathrm{ACR}(X,\mathcal{L}) &\ge \tau_{\mathrm{ACR}}, &
\mathrm{RF}(X,\mathcal{L}) &\ge \tau_{\mathrm{RF}}, \\
\mathrm{LPC}(X,\mathcal{L}) &\ge \tau_{\mathrm{LPC}}, &
\mathrm{GB}(X,\mathcal{L}) &\le \tau_{\mathrm{GB}}, \\
\mathrm{SPDR}(\Pi,\mathcal{L}) &\ge \tau_{\mathrm{SPDR}}, &
\mathrm{AC}(X,\mathcal{L}) &\ge \tau_{\mathrm{AC}}, \\
\mathrm{IS}(\sigma) &\ge \tau_{\mathrm{IS}}.
\end{align*}
\end{definition}

\subsection{Recovery bounds}
\label{app:recovery-bounds}

\paragraph{Remark (post-hoc recovery bounds).}
Recovery operates on the surviving record $\mathcal{L}$ and is bounded by
its content in two ways.

\emph{Field recovery.}
If a field $f$ required by some policy $\pi$ was omitted from every record
entry and left no trace in the surviving content, no post-hoc analysis can
recover it.
Recovery can sometimes infer responsibility-relevant information from
surviving content (e.g., stylistic attribution from output text), but it
cannot materialize fields that are entirely absent from $\mathcal{L}$.

\emph{Integrity.}
Post-hoc signing can raise the current integrity level of a record artifact,
but it cannot retroactively certify that the record was unmodified before the
signing event.
A record that was mutable at write time may have been silently altered before
post-hoc protection was applied.
The integrity guarantee therefore covers only the period after signing, not
the full execution history.

These bounds limit how far the recover mechanism class
(\S\ref{sec:lifecycle}) can compensate for gaps left by detect and enforce.

\section{Platform-Level Security Scan: OpenClaw}
\label{app:openclaw}

The ecosystem evidence in \S\ref{sec:evidence} scans six open-source agent
\emph{projects}. To complement that project-level view, we include a
platform-level scan of OpenClaw~\cite{openclaw2026}, a full-featured open-source AI assistant
with a gateway, extensions, and a skills
marketplace.\footnote{\url{https://github.com/openclaw/openclaw}}
Using agent-audit v0.18.2~\citep{zhang2026agentaudit}, the scan produced 680
raw findings, of which 615 (90.4\%) were auto-classified as likely false
positives---predominantly extension-privilege findings that reflect
OpenClaw's intentional plugin trust model rather than security defects.
After automated triage, 65 findings remained active (31 confirmed, 34
requiring manual review), concentrated in two OWASP categories:

\begin{itemize}[leftmargin=*,itemsep=2pt]
  \item \emph{Credential exposure (ASI-04, 58 findings):} direct macOS
    Keychain access, hardcoded secrets, and NOPASSWD sudoers configuration.
    These reflect OpenClaw's architecture as a personal assistant that
    integrates with system credential stores.
  \item \emph{Code execution (ASI-05, 7 findings):} unsandboxed subprocess
    calls in extensions and a \texttt{curl | bash} pattern in a
    community-contributed skill definition---a classic supply-chain risk.
\end{itemize}

Two findings are particularly relevant to the auditability framework:
\begin{itemize}[leftmargin=*,itemsep=2pt]
  \item A skill instructed the agent to modify \texttt{MEMORY.md}, a file
    that influences long-term agent behavior. A compromised skill or prompt
    injection could use this as a persistence mechanism, poisoning the
    agent's context across future sessions. This is an Evidence Integrity
    concern: the agent's behavioral state can be silently modified without
    detection.
  \item The \texttt{curl | bash} pattern in a skill definition means
    that a dynamically selected skill can trigger arbitrary code execution
    from an unverified external source---an Evidence Integrity risk (code
    runs outside any protected record) that also creates a Responsibility
    Attribution gap, since the resulting actions cannot be traced back
    through the skill's provenance chain.
\end{itemize}

OpenClaw documents a mature trust model that explicitly scopes out several
finding categories (e.g., prompt injection without a boundary bypass,
workspace file writes under operator control). This illustrates the
importance of explicit policy scoping: what counts as a violation depends on
which policies the deployment chooses to enforce, a prerequisite for
meaningful Policy Checkability (\S\ref{subsec:pc}).
The full scan report is available in the agent-audit
repository.


\end{document}